\def\year{2018}\relax
\documentclass[letterpaper]{article} 
\usepackage{aaai18}  
\usepackage{times}  
\usepackage{helvet}  
\usepackage{courier}  
\usepackage{url}  
\usepackage{graphicx}  

\usepackage{caption}
\usepackage{subcaption}
\usepackage{graphicx}
\usepackage{color}
\usepackage{times}
\usepackage{url}
\usepackage{amsmath}
\usepackage{amsthm}
\usepackage{amsfonts}
\usepackage{xspace}
\usepackage{algorithmicx}
\usepackage{algorithm}
\usepackage{algpseudocode}
\usepackage[algo2e]{algorithm2e}
\usepackage{algorithmicx}
\usepackage{algorithm}
\usepackage{algpseudocode}
\usepackage[algo2e]{algorithm2e}

\DeclareMathOperator*{\argmax}{arg\,max}
\DeclareMathOperator*{\argmin}{arg\,min}
\newcommand{\lif}[0]{
  \leftarrow
}
\newcommand{\rif}[0]{
  \rightarrow
}
\newcommand{\vt}[1]{
\mathbf{#1}
}
\newcommand{\fn}[1]{
{\it #1}
}

\newcommand{\xor}[0]{
  \oplus
}
\newcommand{\fzand}[0]{
  \wedge
}
\newcommand{\bAnd}[0]{
  \bigwedge
}
\newcommand{\fzor}[0]{
  \vee
}
\newcommand{\bOr}[0]{
  \bigvee
}

\newcommand{\fziff}[0]{
  \leftrightarrow
}

\newcommand{\pr}[1]{
  \mathrm{#1}
}

\newcommand{\En}[0]{ 
  \fn{E}
}

\newcommand{\fgref}[1]{
  Figure \ref{#1}\xspace
}

\newtheorem{proposition}{Proposition}
\newtheorem{theorem}{Theorem}

\newtheorem{lemma}{Lemma}
\newtheorem{definition}{Definition}
\newtheorem{example}{Example}

\begin{document}
%
\title{Propositional Knowledge Representation and Reasoning in Restricted Boltzmann Machines}
\author{Son N. Tran\\
The Australian E-health research centre, CSIRO\\
Level 5, UQ Health Science Building\\
Brisbane, QLD 4026, Australia\\
}
\maketitle
\begin{abstract}
While knowledge representation and reasoning are considered the keys for human-level artificial intelligence, connectionist networks have been shown successful in a broad range of applications due to their capacity for robust learning and flexible inference under uncertainty. The idea of representing symbolic knowledge in connectionist networks has  been well-received and attracted much attention from research community as this can establish a foundation for  integration of scalable learning and sound reasoning. In previous work,  there exist a number of approaches that map logical inference rules with feed-forward propagation of  artificial neural networks (ANN). However, the discriminative structure of an ANN requires the separation of input/output variables which makes it difficult for general reasoning where any variables should be inferable. Other approaches address this issue by employing generative models such as symmetric connectionist networks, however, they are difficult and convoluted. In this paper we propose a novel method  to represent propositional formulas in restricted Boltzmann machines which is less complex, especially in the cases of logical implications and Horn clauses. An integration system is then developed and evaluated in  real datasets  which shows promising results. 
\end{abstract}
\section{Introduction}

In AI research, there has been a lasting debate over symbolism and
connectionism as they are two key opposed paradigms for information
processing \cite{Smolensky_1987,Minsky_1991}. The former has been
known as the foundation language of AI which captures higher level of
intelligence with explainable and reasoning capability. The latter is
getting more attention  due to its indisputable advantages in
scalable learning and dealing with noisy data. Despite their difference, there
is a strong argument that combination of the two should offer joint
benefits \cite{Smolensky_1995,Valiant_2006,Garcez_2008}.
This was the major motivation for  consistent efforts in developing  integration
systems in the last two decades  \cite{Towel_1994,Garcez_1999,Leo_2011,Franca_2014,Son_2016}. Such
systems are well known not only for better performance in making decision but also for more
efficient learning. The key success here is lying on a mechanism to
represent symbolic knowledge in a connectionist network.  This is also
useful for knowledge extraction \cite{Towell_1993,Garcez_2001}, i.e. to
seek for symbolic representation of the networks.

In previous work artificial neural networks (ANNs) play a central role
in encoding symbolic knowledge. ANNs are black-boxes of input-output
mapping function which would be more transparent when their parameters
are constrained to perform logical inference, such as
modus ponens \cite{Towel_1994,Garcez_1999,Franca_2014}.  However, due
to the discriminative structure of such ANNs only a subset of
variables, i.e. the consequences in {\it if-then formulas}, can be
inferred as outputs, while the other variables, i.e. the antecedents,
are encoded as inputs. This would not represent the behaviour of
logical formulas and therefore cannot support general reasoning where
any variables is inferable.  In order to solve this issue generative
neural networks should be employed as they can treat all variables on
a non-discriminatory basis, hence are more useful for symbolic
reasoning.  In Penalty logic \cite{Pinkas_1995}, it has been shown
that well-formed formulas can be represented by a symmetric
connectionist network (SCN) where the lower energy of the network
corresponds to the less violation of the formulas. However, learning
and inference in SCNs are difficult due to the intractability
problem. Interestingly, generative neural networks with latent
variables such as restricted Boltzmann machines
(RBMs) \cite{Smolensky_1986}, a simplified variant of SCNs, can learn
to represent semantic patterns from large amounts of data
efficiently \cite{Hinton_2002,Lee_2007}. Besides, inference with RBMs
is also less complex than other SCNs with dense connections such as
Boltzmann machines (BMs), because in RBMs connections between units in the
same layers are discarded. Therefore, representation of symbolic
knowledge in RBMs would offer an advantage of practicality.

Several attempts have been made recently to integrate symbolic
representation and RBMs \cite{Leo_2011,Son_2016}. Despite
achieving good practical results they are still heuristic and lack a
supporting theory. The most related literature to this work is Penalty
logic \cite{Pinkas_1995} as we already mentioned above.  However, it
is difficult to apply Penalty logic to restricted Boltzmann
machines. In this paper we propose a new method to represent
propositional formulas in RBMs where symbolic reasoning can be seen as
minimising the network's energy function. The idea is to convert a
formula into a disjunctive normal form (DNF) in which at most one conjunctive
clause holds given an assignment, and then apply variable
elimination to achieve a more compact DNF. For logical implications and
Horn clauses, both popular logical forms for knowledge bases in practice, this
conversion is efficient.  We then apply the theoretical result to
develop CRILP-Confidence rule inductive logic programming system for
encoding, learning and inference with symbolic knowledge using RBMs. In the
experiments we show that a learned RBM can encapsulate  symbolic knowledge in
the form of conjunctive clauses. We also show that our CRILP gives
promising results in comparison with other neural-symbolic systems
such as CILP \cite{Garcez_1999} and CILP++ \cite{Franca_2014}, and with
the state-of-the-art inductive logic programming system
Aleph \cite{aleph}.

The paper is organised as follows: in the next section we review
related work on knowledge representation in connectionist networks and
graphical models. After that we propose a method to transform and
convert propositional formulas to RBMs' energy functions. We then show
the efficiency of the method when applied to logical implications, and
we also show a conceptual comparison to other methods for representing
knowledge in RBMs.  Next, we apply the theoretical result to build a
model that encodes knowledge into RBMs for tuning and reasoning. In
the experiment section we validate the proposed method and model on
real datasets. Finally, in the last section we conclude the paper.

\section{Background and Related Work}
\label{sec:background}

In artificial neural networks, symbolic knowledge representation is based on the equivalence between feed-forward inference of the networks and modus ponens of logical rules. One of the earliest work is Knowledge-based artificial neural network \cite{Towel_1994} which encodes {\it if-then} rules in a hierarchy of perceptrons. In another approach \cite{Garcez_2001} an 1-hidden layer neural networks with recurrent connections is proposed to support more complex rules. An extension of this system, called CILP++, uses {\it bottom clause propositionalisation} technique to work with first-order logic \cite{Franca_2014}. Logic tensor network \cite{Serafini_2016} employs neural embedding to transform symbols to a vector space where logical inference is carried out through matrix/tensor computation. 

Symbolic representation in graphical models has also been commonly
studied. For example, in a notable work \cite{Richardson_2006} Markov networks are
employed to generalise first-order logic. This work is different from
ours in that it combines statistical and logical inference
while we show the relation between the former and the latter. Besides the logical
knowledge we study here, some other work also show the advantage of
learning structural knowledge in graphical models, especially for
tractable inference \cite{Poon2011,Darwiche_2003}.

In this work, we focus on restricted Boltzmann machines, a class of
connectionist networks with interconnected units similar to neural
networks but for presenting a probabilistic distribution.  An RBM can be
seen as a two-layer neural  network with bidirectional connections
which is characterised by a function called energy:
{
\begin{equation}
\label{eq:rbm_en}
  \En(\vt{x},\vt{h}) = -\sum_{i,j} w_{ij}x_ih_j - 
\sum_{i}a_ix_i - \sum_jb_jh_j
\end{equation}
}
where $a_i$ and $b_j$ are biases of $x_i$, $h_j$ respectively; $w_{ij}$ is the connection weight between $x_i$ and $h_j$. This RBM presents a joint distribution  $p(\vt{x},\vt{h})= \frac{1}{Z}e^{-\frac{1}{\tau}\En(\vt{x},\vt{h})}$ where $Z=\sum_{\vt{x}\vt{h}}e^{-\frac{1}{\tau}\En(\vt{x},\vt{h})}$ is the partition function, and $T$ is the temperature.

Penalty logic is among the earliest attempts to show that any propositional formulas can be represented in symmetric connectionist systems,  which can also be applied to RBMs \cite{Pinkas_1995}.  Penalty logic
explains the relation between propositional formulas and SCNs. 
Penalty logic formulas are defined as a finite set of pairs
$(\rho,\varphi)$, in which each propositional well-formed formula (WFF)
$\varphi$ is associated with a real value $\rho$ called {\it
penalty}. A violation rank function $V_{rank}$ is defined as
the sum  of the penalties from violated formulas.  A preferred
model is a truth-value assignment $\vt{x}$ that has minimum
total penalty. Applied to classification, for example, to decide the
truth-value of a target proposition $\pr{y}$ given an assignment
$\pr{\vt{x}}$ of the other propositions, one will choose the value of
$\pr{y}$ that minimises $V_{rank}(\pr{\vt{x}},\pr{y})$. 
Reasoning with Penalty logic is shown to be equivalent to minimising
an energy function in a SCN \cite{Pinkas_1995}. This is the fundamental to form a link
between propositional logic program and the network.

The Penalty logic idea seems to work straightforwardly with dense structures such as higher-order Boltzmann machines, however it is computationally expensive to represent a formula in RBMs, despite that compared to BMs learning in RBMs is easier due to the efficient inference mechanism. More importantly, it shows that by stacking
several RBMs, one on top of another we can not only extract different
level of abstractions from domain's data but also achieve better
performance \cite{Hinton_2006,Lee_2007}. Recently,
several attempts have been made to extract and encode symbolic
knowledge into RBMs \cite{Leo_2011,Son_2016}. However, it is not
theoretically clear how such knowledge is represented in the RBMs formally.

 
From a statistical perspective, representing a propositional formula is similar to representing a uniform distribution over all assignments that satisfy the formula, i.e. make the formula hold. In this paper such assignments are referred to as {\it preferred models}. Since RBMs are universal approximators, it is true that any discrete distribution can be ``approximated arbitrarily well'' \cite{LeRoux_2008}, and therefore we can always find an RBM to represent a uniform distribution of preferred models. However, while that work utilises statistical methods over the preferred models of formulas resulting in a very large network, our work focuses on logical calculus to transform and convert formulas directly to the energy function of a more compact RBM.

\section{Theoretical Study}
\label{theory}
\subsection{Propositional Calculus and RBMs}
\label{dnf_rbm}
Since symmetric connectionist networks are a generalisation of RBMs we
can apply Penalty logic to represent propositional knowledge in that
restricted variant. However, it is unnecessarily complicated that,
according to the proposed algorithms in \cite{Pinkas_1995}, we need to
construct a higher-order BM and then transform its energy function
into a quadratic form by adding more hidden variables. The former step
is computationally expensive while the latter step requires a large
number of hidden variables to be added until every energy term has at
most one visible variable. The complexity for the construction of a
high-order energy function in the first step can be reduced by
converting a formula into a conjunction of sub-formulas, each of at
most three variables. However this would need more hidden variables
and not always results in an RBM since there can be more than one
hidden variables in an energy term. This paper introduces a much
simpler method to represent a well-formed formula in an RBM, and
extends it to represent a logical program. We show that this could be
accomplished by converting WFFs into disjunctive normal form, as
detailed below, instead of conjunctions of sub-formulas as in Penalty
logic.

In propositional logic, any WFF $\varphi$ can be represented in disjunctive
normal form (DNF) \cite{Stuart_2003_part3}:
{
\begin{equation*}
\varphi \equiv \bOr_j (\bAnd_{t \in \mathcal{S}_{T_j}} \pr{x}_t \fzand \bAnd_{k \in \mathcal{S}_{K_j}} \neg \pr{x}_{k})
\end{equation*}
}
where each $(\bAnd_{t \in \mathcal{S}_{T_j}} \pr{x}_t \fzand \bAnd_{k \in
\mathcal{S}_{K_j}} \neg \pr{x}_{k})$ is called a ``conjunctive clause''. Here we
denote the literals as $\pr{x}_t$, $\pr{x}_k$ and $\mathcal{S}_{T_j}$ and $\mathcal{S}_{K_j}$ are
the set of $T_j$ indices of positive literals and the set of $K_j$ indices of negative literals
respectively.

\begin{definition}$\text{ }$
  \begin{itemize} \item A ``strict DNF'' (SDNF) is a DNF where there is at most
    one single conjunctive clause is $True$ at a time.  \item A ``full DNF'' is
    a DNF where each variable must appear at least once in every
    conjunctive clause.  \end{itemize}
\end{definition}

Any propositional well-formed formula can be presented as a
SDNF. Indeed, for example, suppose that $\varphi$ is a WFF in disjunctive normal
form. If $\varphi$ is not SDNF then there exist some groups of
conjunctive clauses which are $True$ given a preferred assignment. We
can always convert this group of conjunctive clauses to a full DNF
which is also a SDNF.


\begin{definition}
A WFF $\varphi$ is equivalent to a neural network $\mathcal{N}$ if and only if for any
truth assignment $\vt{x}$, $s_\varphi(\vt{x}) = - A\En_{rank}(\vt{x}) + B$,
where $s_\varphi(\vt{x}) \in \{0,1\}$ is the truth value of $\varphi$ given
$\vt{x}$ with $True\equiv 1$ and $False \equiv 0$; $A>0$ and $B$ are constants; $\En_{rank}(\vt{x})
= min_\vt{h}\En(\vt{x},\vt{h})$ is the energy ranking function of
$\mathcal{N}$ minimised over all hidden units.
\end{definition} 
This definition of equivalence is similar to that of
 Penalty logic \cite{Pinkas_1995}. Here the equivalence guarantees that all preferred
 models of a WFF $\varphi$ would also minimise the energy of the  network $\mathcal{N}$. 

\begin{lemma}
\label{lem:wff2ecs} Any SDNF $\varphi \equiv \bOr_j (\bAnd_{t \in \mathcal{S}_{T_j}} \pr{x}_t \fzand \bAnd_{k \in \mathcal{S}_{K_j}} \neg \pr{x}_{k})$ can be mapped onto a SCN with energy function $\En = -\sum_{j} \prod_{t \in \mathcal{S}_{T_j}} x_t \prod_{k \in \mathcal{S}_{K_j}} (1-x_{k}) $ where $\mathcal{S}_{T_j}$, $\mathcal{S}_{K_{j}}$ are the set of $T_j$ indices of positive literals and the set of $K_j$ indices of negative literals
respectively.  
\end{lemma}
\begin{proof}
By definition, $\varphi \equiv \bOr_j (\bAnd_{t\in \mathcal{S}_{T_j}} \pr{x}_t \fzand
\bAnd_{k \in \mathcal{S}_{K_j}} \neg \pr{x}_{k})$. Each conjunctive clause
$\bAnd_{t\in \mathcal{S}_{T_j}} \pr{x}_t \fzand \bAnd_{k \in \mathcal{S}_{K_j}} \neg \pr{x}_{k}$
corresponds to $\prod_{t\in \mathcal{S}_{T_j}} x_t \prod_{k\in \mathcal{S}_{K_j}} (1-x_{k})$ which
maps to $1$ if and only if $x_t=1$ ($\pr{x}_t=True$) and $x_{k}=0$
($\pr{x}_{k}=False$) for all $t \in \mathcal{S}_{T_j}$ and $k \in \mathcal{S}_{K_j}$. Since
$\varphi$ is a SDNF, it is $True$ if and only if one conjunctive
clause is $True$, then the sum $\sum_{j} \prod_{t\in \mathcal{S}_{T_j}} x_t
\prod_{k \in \mathcal{S}_{K_j}} (1-x_{k})=1$ if and only if the assignment of
truth-values for $x_t$ , $x_{k}$ is a preferred model of
$\varphi$. Hence, there exists a SCN with energy
function $\En=-\sum_{j} \prod_{t\in \mathcal{S}_{T_j}} x_t \prod_{k \in \mathcal{S}_{K_j}}
(1-x_k)$ such that $s_\varphi(\vt{x}) = - \En_{rank}(\vt{x})$.
\end{proof}

We now show that any formula can be converted into RBMs by using its SDNF form.

\begin{theorem}
\label{theorem:prop_rbm} 
 Any SDNF $\varphi \equiv \bOr_j (\bAnd_{t \in \mathcal{S}_{T_j}} \pr{x}_t \fzand \bAnd_{k \in \mathcal{S}_{K_j}} \neg \pr{x}_{k})$ can be mapped onto an equivalent RBM with energy function $\En = -\sum_jh_j(\sum_{t \in \mathcal{S}_{T_j}} x_t - \sum_{k \in \mathcal{S}_{K_j}}x_{k}  -  T_j+ \epsilon)$, where $0<\epsilon<1$;  $\mathcal{S}_{T_j}$, $\mathcal{S}_{K_{j}}$ are the set of $T_j$ indices of positive literals and the set of $K_j$ indices of negative literals
respectively.
\end{theorem}

\begin{proof} 
We have seen in Lemma \ref{lem:wff2ecs} that any SDNF $\varphi$ can be
mapped onto energy function $\En = -\sum_{j} \prod_{t\in \mathcal{S}_{T_j}}
x_t \prod_{k \in \mathcal{S}_{K_j}} (1-x_{k}) $. Let us denote $T_j$ as the number
of positive propositions in a conjunctive clause $j$. For each term
$\tilde{e}_j(\vt{x}) = -\prod_{t \in \mathcal{S}_{T_j}} x_t \prod_{k \in \mathcal{S}_{K_j}}
(1-x_{k})$ we define an energy term with an hidden variable
$h_j$ as: $e_j(\vt{x},h_j) = -h_j(\sum_{t \in \mathcal{S}_{T_j}} x_t
- \sum_{k \in \mathcal{S}_{K_j}}x_{k} -T_j + \epsilon) $ with $0< \epsilon < 1$ such that
$\tilde{e}_j(\vt{x}) = \frac{e_{j \text{ }
rank}(\vt{x})}{\epsilon}$, where $e_{j \text{ }
rank}(\vt{x})=min_{h_j}e_j(\vt{x},h_j)$. This equation holds because $-(\sum_{t \in \mathcal{S}_{T_j}}
x_t - \sum_{k \in \mathcal{S}_{K_j}} x_{k} -T_j+\epsilon) = -\epsilon$ if and only if $x_t=1$ and
$x_{k}=0$ for all $t \in \mathcal{S}_{T_j}$ and $k \in \mathcal{S}_{K_j}$, which makes $min_{h_j}
e_j(\vt{x},h_j)=-\epsilon$ with $h_j=1$. Otherwise $-(\sum_t x_{t \in \mathcal{S}_{T_j}}
- \sum_{k \in \mathcal{S}_{K_j}} x_{k} -T_j+\epsilon) >0$ and then $min_{h_j} e_j(\vt{x},h_j) =
0$ with $h_j=0$. By repeating the process on every term
$\tilde{e}_j(\vt{x})$ we can conclude that any SDNF $\varphi$ is
equivalent with an RBM having the energy function:
{
 \begin{equation}
   \label{eq:prop_rbm_en}
   \En = -\sum_jh_j(\sum_{t \in \mathcal{S}_{T_j}} x_t - \sum_{k \in \mathcal{S}_{K_j}}x_{k}  -  T_j+ \epsilon)  
 \end{equation}
 }
where:  $s_\varphi(\vt{x}) = -\frac{1}{\epsilon} \En_{rank}(\vt{x})$
\end{proof}

\begin{example}
The XOR formula $(\pr{x} \xor \pr{y}) \fziff \pr{z}$ can be converted into a SDNF as:
{
\begin{equation*}
\varphi \equiv (\neg \pr{x} \fzand \neg \pr{y} \fzand \neg \pr{z}) \fzor (\neg \pr{x}
\fzand \pr{y} \fzand \pr{z}) \fzor (\pr{x} \fzand \neg \pr{y} \fzand \pr{z}) \fzor (\pr{x} \fzand \pr{y}
\fzand \neg \pr{z})
\end{equation*}
}
For each conjunctive clause, for example $\pr{x} \fzand \pr{y} \fzand \neg \pr{z}$ we
create a term $xy(1-z)$ and add it to the energy function. After all
terms are added, we have the energy function of a SCN as:
\begin{equation*}
\begin{aligned}
 \En = &-(1-x)(1-y)(1-z) - xy(1-z) \\
       &-x(1-y)z - (1-x)yz
\end{aligned}          
\end{equation*}
\begin{figure}[ht]
\centering
\includegraphics[width=.4\textwidth]{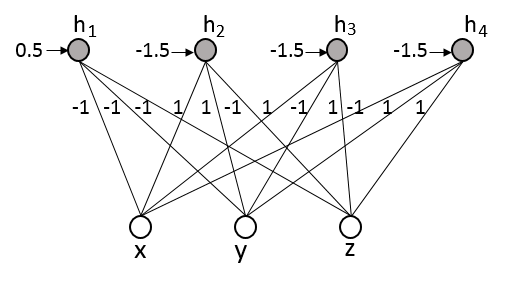}
\caption{RBM for XOR using DNF}
\label{dnf_xor}
\caption{RBMs for XOR formula: $(\pr{x} \xor \pr{y}) \fziff \pr{z}$}
\label{fig:xor2rbm}
\end{figure}
Applying Theorem \ref{theorem:prop_rbm} we can construct an RBM for a
XOR function as shown in
\fgref{dnf_xor}. In this example we choose $\epsilon=0.5$. The energy function of this RBM is:
{
\begin{equation*} 
  \begin{aligned} \En &= -h_1(-x - y - z + 0.5) - h_2(x + y -  z - 1.5)\\
  & - h_3(x+y-z-1.5) - h_4( -x + y + z - 1.5)
  \end{aligned}
\end{equation*}
}

In Table \ref{tab:en_xor_equip}, we show the equivalence between $\En_{rank}$  and the truth values of the XOR.
\begin{table}[ht]
    \centering
    {\scriptsize
    \begin{tabular}{|c|c|c||c|c|}
      \hline
      \hline
      $\pr{x}$     &   $\pr{y}$   & $\pr{z}$     & $s_\varphi(\pr{x},\pr{y},\pr{z})$  &  $\En_{rank}$ \\
      \hline
      $0$ & $0$ & $0$ & $True$              &  $-0.5$\\
      \hline
      $0$ & $0$ & $1$ & $False$             &  $0$\\
      \hline
      $0$ & $1$ & $0$ & $False$             &  $0$\\
      \hline
      $0$ & $1$ & $1$ & $True$              &  $-0.5$\\
      \hline
      $1$ & $0$ & $0$ & $False$             &  $0$\\
      \hline
      $1$ & $0$ & $1$  & $True$              &  $-0.5$\\
      \hline
      $1$ & $1$ & $0$ & $True$              &  $-0.5$\\
      \hline
      $1$ & $1$ & $1$  &$False$             &  $0$\\
      \hline
    \end{tabular}
    }
    \caption{Minimised energy function of RBM and truth table of XOR formula}
    \label{tab:en_xor_equip}
  \end{table}
\end{example}

\subsection{Propositional logic program}
We can present a set of formulas $\Phi = \{\varphi_1, ...\varphi_N\}$ in an RBM by applying Theorem \ref{theorem:prop_rbm} to each formula
$\varphi_i$. Similarly, for a set of weighted
formulas $\Phi = \{w'_1: \varphi_1, ..., w'_N:\varphi_N\}$  we can also construct an equivalent RBM where each energy term generated from formula $\varphi_i$ is multiplied with its weight $w'_i$. In both cases, given an assignment, the minimum energy of the RBMs is also the maximum satisfiability of the logic program $\Phi$, i.e. the weighted sum of the formulas being $True$ given the assignment.  
\begin{proposition}
  \label{prop:rbm_lprogram} For a weighted knowledge base
$\Phi=\{w'_1:\varphi_1,...,w'_N:\varphi_N\}$ there exists an
equivalent RBM $\mathcal{N}$  such that
$s_\Phi(\vt{x}) = -\frac{1}{\epsilon}\En_{rank}(\vt{x})$, where $\En_{rank}(\vt{x})$ is the energy ranking function of $\mathcal{N}$ and $s_\Phi(\vt{x})$ is the weighted satisfiability given a truth assignment $\vt{x}$.
\end{proposition}

\begin{proof}
A formula $\varphi_i$ can be decomposed into a set of weighted conjunctive clauses from its SDNF. Same conjunctive clauses from the program $\Phi$ can be combined by summing their weights. If there exist two conjunctive  clauses and one is subsumed by the other, then the former can be removed while the other's weight is replaced by the sum of their weights. We call the weights of the conjunctive clauses  ``confidence-values'' ($c$) to distinguish them from the weights of the formulas ($w'$). These steps would create a set of generalised and unique weighted conjunctive clauses. Now, from Theorem \ref{theorem:prop_rbm}  we know that a conjunctive clause
$\bAnd_{t \in \mathcal{S}_{T_j}}\pr{x}_t \fzand \bAnd_{k \in
\mathcal{S}_{K_j}} \neg \pr{x}_{k}$ is equivalent to an energy term
$e_j(\vt{x},h_j) = -h_j(\sum_{t\in \mathcal{S}_{T_j}} x_t
- \sum_{k \in \mathcal{S}_{K_j}}x_{k} -|T_j|+ \epsilon)$ with $0<\epsilon<1$. A weighted conjunctive clause $c_j: \bAnd_{t \in \mathcal{S}_{T_j}}\pr{x}_t \fzand \bAnd_{k \in
\mathcal{S}_{K_j}} \neg \pr{x}_{k}$, therefore, is equivalent to an energy term
 $c_j\times e_j(\vt{x},h_j)$. For each weighted conjunctive clause we can add a hidden unit to an RBM and assign the connection weights as
 $w_{tj} = c_j$ and $w_{kj}=-c_j$ for all $t\in \mathcal{S}_{T_j}$ and $k \in
 \mathcal{S}_{K_j}$. The bias for this hidden unit is $c_j(-|T_j|+\epsilon)$. The weighted knowledge base and the RBM is equivalent such that $s_\Phi
 = -\frac{1}{\epsilon} E_{rank}$\footnote{$s_\Phi$ in this case is the
 sum of the weights from satisfied rules}.
\end{proof}
\begin{example}  Nixon diamond.
{\footnotesize
\begin{align*}
&1000: \pr{n} \rif \pr{r}  \quad \text{ Nixon is a republican.} \\       
&1000: \pr{n} \rif \pr{q}  \quad \text{ Nixon is also a quaker.}\\
&10\text{ }\text{ }\text{ }  : \pr{r} \rif \neg \pr{p} \quad \text{republicans tend not to be pacifist.}\\
&10\text{ }\text{ }\text{ }  : \pr{q} \rif \pr{p} \quad \text{quakers tend to be pacifist.}
\end{align*}
}
\begin{figure}[ht]
\centering
\includegraphics[width=0.3\textwidth]{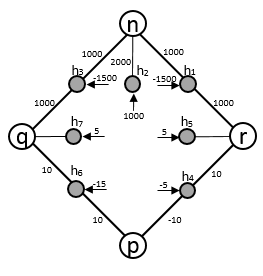}
\caption{The RBM for Nixon diamond program.}
\label{fig:rbm_nixon}
\end{figure}
Let us convert all fomulas to  SDNFs, for example $\pr{n} \rif \pr{r} \equiv (\pr{n} \fzand \pr{r})\fzor (\neg \pr{n})$, which results in $8$ conjunctive clauses. After combining the weights of clause $\neg \pr{n}$ which appears twice, we create an RBM from the following unique conjunctive clauses and their confidence values:
\begin{equation*}
\begin{aligned}
1000&: \pr{n} \fzand \pr{r}, \quad 2000: \neg \pr{n}, \quad 1000: \pr{n} \fzand \pr{q}\\
10&: \pr{r} \fzand \neg \pr{p}, \quad 10: \neg \pr{r}, \quad 10: \pr{q} \fzand \pr{p}, \quad 10: \neg \pr{q}\\
\end{aligned}
\end{equation*}
The final RBM should have an energy function as following, with $\epsilon=0.5$:
{
\begin{align*}
\En = &-h_1(1000n+1000r-1500) - h_2(-2000n+1000)\\
    &-h_3(1000n + 1000q-1500)-h_4(10r - 10p - 5)\\
    &-h_5(-10r + 5) -h_6(10q + 10p - 15)\\
    &- h_7(-10q + 5)
\end{align*}
}
and its graphical structure is shown in Figure \ref{fig:rbm_nixon}.

\end{example}
\subsection{Reasoning and Sampling}
We now show a relation between inference in RBMs and maximising satisfiability.
\begin{proposition}
  \label{prop:gibbs_sat}
Let $\mathcal{N}$ be an RBM constructed from a weighted knowledge base $\Phi$; $\mathcal{A}$ be a set of indices of assigned  variables $\vt{x}_\mathcal{A} = \{x_\alpha|\alpha \in \mathcal{A}\}$; $\mathcal{B}$ is the set of indices of the rest, unassigned variables $\vt{x}_\mathcal{B} = \{x_\beta|\beta \in \mathcal{B}\}$, inference of $\vt{x}_\mathcal{B}$ given $\mathcal{N}$ and $\vt{x}_\mathcal{A}$ with Gibbs sampling is equivalent to a search for $\vt{x}_\mathcal{B}$ to maximise the satisfiability of $\Phi$.
\end{proposition}
\begin{proof}
 It has been shown in Proposition \ref{prop:rbm_lprogram} that the satisfiability is inversely proportional to the $\En_{rank}$ function of the RBM, such that:
 \begin{equation}
   s_\Phi(\vt{x}_\mathcal{B},\vt{x}_\mathcal{A}) \propto -\En_{rank}(\vt{x}_\mathcal{B},\vt{x}_\mathcal{A})
 \end{equation}
 where $\En_{rank}(\vt{x}_\mathcal{B},\vt{x}_\mathcal{A}) = \min_\vt{h} \En(\vt{x}_\mathcal{B},\vt{x}_\mathcal{A},\vt{h})$. Therefore, a value of $\vt{x}_\mathcal{B}$ that minimises the energy function also maximises the satisfiability because:
 \begin{equation}
 \begin{aligned}
 \vt{x}_\mathcal{B}^* &= \argmin_{\vt{x}_\mathcal{B}} \min_\vt{h}\En(\vt{x}_\mathcal{B},\vt{x}_\mathcal{A},\vt{h}) \\
 & = \argmin_{\vt{x}_\mathcal{B}} -s_\Phi(\vt{x}_\mathcal{B},\vt{x}_\mathcal{A})\\ 
 &= \argmax_{\vt{x}_\mathcal{B}} s_\Phi(\vt{x}_\mathcal{B},\vt{x}_\mathcal{A})
 \end{aligned}
 \end{equation}

 Now, let us consider an iterative process to search for $\vt{x}_\mathcal{B}^*$ by minimising the energy function. This can be done by using gradient descent to alternatively update the values of $\vt{h}$ and $\vt{x}_\mathcal{B}$  one at a time to minimise $\En(\vt{x}_\mathcal{B},\vt{x}_\mathcal{A},\vt{h})$ while keeping the other variables fixed. This update can be repeated until convergence. Note that the gradients:
 \begin{equation}
   \label{eq:en_grad}
   \begin{aligned}
     \frac{\partial -\En(\vt{x}_\mathcal{B},\vt{x}_\mathcal{A},\vt{h})}{\partial h_j} &= \sum_{i\in \mathcal{A}\cup\mathcal{B}} x_iw_{ij} + b_j\\
     \frac{\partial -\En(\vt{x}_\mathcal{B},\vt{x}_\mathcal{A},\vt{h})}{\partial x_\beta} &= \sum_j h_jw_{\beta j} + a_\beta
    \end{aligned}
 \end{equation}

 In the case of Gibbs sampling, given the assigned variables $\vt{x}_\mathcal{A}$ the process starts with an initialisation of $\vt{x}_\mathcal{B}$,  and then alternatively infer the hidden variables $h_j$ and the unassigned visible variables $x_\beta$  using the conditional distributions $h_j \sim p(h_j|\vt{x})$ and $x_\beta \sim p(x_\beta|\vt{h})$ respectively, where $\vt{x}=\{\vt{x}_\mathcal{A},\vt{x}_\mathcal{B}\}$ and 
 \begin{equation}
   \label{eq:gibbs}
   \begin{aligned}
     p(h_j|\vt{x}) & = \frac{1}{1+e^{-\frac{1}{\tau}\sum_i x_i w_{ij}+b_j}}  = \frac{1}{1+e^{-\frac{1}{\tau} \frac{\partial -\En(\vt{x}_\mathcal{B},\vt{x}_\mathcal{A},\vt{h})}{\partial h_j}}}\\
     p(x_\beta|\vt{h}) & = \frac{1}{1+e^{-\frac{1}{\tau}\sum_j h_j w_{\beta j}+a_\beta}}  =  \frac{1}{1+e^{-\frac{1}{\tau} \frac{\partial -\En(\vt{x}_\mathcal{B},\vt{x}_\mathcal{A},\vt{h})}{\partial x_\beta}}}\\
   \end{aligned}
  \end{equation}
It can be seen from \eqref{eq:gibbs} that the distributions are monotonic functions of  the negative energy's gradient over $\vt{h}$ and $\vt{x}_\mathcal{B}$. Therefore, performing Gibbs sampling from those functions can be seen  as moving randomly towards a local point of minimum energy, or equivalently to a maximum satisfiability. In the case of $\tau=0$, this process becomes deterministic.
\end{proof}
Intuitively, the energy function of the RBM and the satisfiability of the knowledge base is inversely proportional, therefore every step of Gibbs sampling to reduce the energy function will also increase the satisfiability.
\section{Logical Implication and Horn Clauses}
\label{subsec:logic_implication}
\subsection{Representation}
In previous section we showed that any propositional logic program can be equivalently converted into an RBM. In general cases, transforming an arbitrary formula to disjunctive normal form is exponentially hard. Fortunately, in practice many knowledge bases are in the form of logical implications with which the conversion can be efficient. Let us consider a general presentation of a logical implication $\varphi$ as:
{
\begin{equation}
\label{horn}
\varphi \equiv \pr{y} \lif \bAnd_{t\in \mathcal{S}_T} \pr{x}_{t} \fzand \bAnd_{k \in \mathcal{S}_K} \neg \pr{x}_k
\end{equation}
}
Here, for ease of presentation in this section we denote the sets of the indices of positive and negative propositions as $\mathcal{S}_T$ and $\mathcal{S}_k$ respectively. Note that if $\mathcal{S}_K$ is empty ($K=0$) this formula is exactly a Horn clause. A SDNF form of this formula can be constructed by creating a conjunctive clause from each preferred model, which would make the conversion complexity exponentially expensive and require $2^{K+T+1}-1$ hidden units to construct an RBM. Fortunately, we can show that it only needs $K+T$ hidden units to represent a logical implication in an RBM.
\begin{theorem}
\label{theorem:horn}
A logical implication $\pr{y} \lif \bAnd_{t\in \mathcal{S}_T} \pr{x}_{t} \fzand \bAnd_{k \in \mathcal{S}_K} \neg \pr{x}_k$, with $\mathcal{S}_{T}$, $\mathcal{S}_{K}$ are respectively the sets of positive and negative propositions' indices, can be represented by an RBM with the energy function: {\footnotesize 
\begin{equation}
\label{sdnf_rbm}
\begin{aligned}
&E = -h_y(\sum_{t\in\mathcal{S}_T}x_t - \sum_{k\in\mathcal{S}_K}x_k +y - T-1+\epsilon)\\
&-\sum_{p \in \mathcal{S}_T \cup \mathcal{S}_K} h_p(\sum_{t\in\mathcal{S}_T.\backslash p}x_t - \sum_{k\in\mathcal{S}_K.\backslash p}x_k + x'_p - |\mathcal{S}_T.\backslash p| - \mathbb{I}_{p\in\mathcal{S}_K} +\epsilon)
\end{aligned}
\end{equation}
}
where 
$\mathcal{S} .\backslash p$ denotes a set $\mathcal{S}$ with $p$ being removed and  $|\mathcal{S}.\backslash p|$ is the size of that set; $x'_p=-x_p$ and $\mathbb{I}_{p\in\mathcal{S}_K}=0$ if $p \in \mathcal{S}_T$, otherwise $x'_p = x_p$  and $\mathbb{I}_{p\in\mathcal{S}_K}=1$.
\end{theorem}
\begin{proof}
A logical implication $\pr{y} \lif \bAnd_{t\in \mathcal{S}_T} \pr{x}_{t} \fzand \bAnd_{k \in \mathcal{S}_K} \neg \pr{x}_k$ can be transformed to a disjunctive normal form as:

{\footnotesize 
\begin{equation}
\label{logimp_dnf}
(\pr{y} \fzand \bAnd_{t\in \mathcal{S}_T} \pr{x}_t \fzand \bAnd_{k\in \mathcal{S}_K} \neg \pr{x}_k) \fzor (\bOr_{t\in \mathcal{S}_T} \neg \pr{x}_t \fzor \bOr_{\in \mathcal{S}_K} \pr{x}_k)
\end{equation}
}
Here, the logical implication holds if and only if either the conjunctive clause in \eqref{logimp_dnf} holds or the disjunctive clause holds. Let us consider the disjunctive clause.
{\footnotesize
\begin{equation}
\gamma \equiv \bOr_{t\in \mathcal{S}_T} \neg \pr{x}_t \fzor \bOr_{\in \mathcal{S}_K} \pr{x}_k
\end{equation}
}
which can be presented as $\gamma \equiv \gamma' \fzor \pr{x}'$, where $\pr{x}'$ can be either $\neg \pr{x}_t$ or $\pr{x}_k$ for any $t \in \mathcal{S}_T$ and $k \in \mathcal{S}_K$; $\gamma'$ is a disjunctive clause obtained by removing $\pr{x}'$ from $\gamma$.  We have: 
\begin{equation}
\label{conj_ext}
\gamma  \equiv (\neg \gamma' \fzand \pr{x}') \fzor \gamma'
\end{equation}
because $(\neg \gamma' \fzand \pr{x}') \fzor \gamma' \equiv (\gamma' \fzor \neg \gamma') \fzand (\gamma' \fzor \pr{x}') \equiv True \fzand (\gamma' \fzor \pr{x}')$. Note that negation of a disjunction is a conjunction, e.g. $\neg(\neg \pr{x}_t\fzor \pr{x}_k)\equiv  \pr{x}_t\fzand\neg\pr{x}_k$,  therefore we can convert $\neg \gamma' \fzand \pr{x}'$ into a conjunctive clause. By applying \eqref{conj_ext}, each time we can eliminate a variable out of a disjunctive clause by moving it to a new conjunctive clause. We can see that the disjunctive clause $\gamma$ holds if and only if either the  disjunctive clause $\gamma'$ holds or the conjunctive clause ($\neg \gamma' \fzand \pr{x}'$) holds. Therefore, at the end, the original disjunctive clause $\gamma$ is transformed into a SDNF. Combine with \eqref{logimp_dnf} we have the SDNF of the logical implication as:
{
\begin{equation}
\label{short_horn_dnf}
\begin{aligned}
&(\pr{y} \fzand \bAnd_{t\in \mathcal{S}_T}  \pr{x}_t  \bAnd_{k\in \mathcal{S}_K} \neg \pr{x}_k) \\
& \fzor \bOr_{p \in \mathcal{S}_T \cup \mathcal{S}_K} (\bAnd_{t\in\mathcal{S}_T.\backslash p} \pr{x}_t \fzand  \bAnd_{k\in\mathcal{S}_K.\backslash p} \neg\pr{x}_k \fzand \pr{x}'_{p})
\end{aligned}
\end{equation}
 }
 where $\mathcal{S} .\backslash p$ denotes a set $\mathcal{S}$ with $p$ has been removed; $\pr{x}'_p\equiv\neg\pr{x}_p$ if $p\in\mathcal{S}_T$ otherwise $\pr{x}'_p\equiv\pr{x}_p$.
Apply Theorem \ref{theorem:prop_rbm}, the energy function of an RBM constructed from this SDNF is detailed as following.
\begin{itemize}
\item  For the disjunctive clause $\pr{y} \fzand \bAnd_{t\in \mathcal{S}_T} \fzand \pr{x}_t  \bAnd_{k\in \mathcal{S}_K} \neg \pr{x}_k$ we create an energy term $-h_y(\sum_{t\in\mathcal{S}_T}x_t - \sum_{k\in\mathcal{S}_K}x_k +y - T-1+\epsilon)$.

\item For each disjunctive clause $\bAnd_{t\in\mathcal{S}_T.\backslash p} \pr{x}_t \fzand  \bAnd_{k\in\mathcal{S}_K.\backslash p}^K \neg\pr{x}_k \fzand \pr{x}'_{p}$ we create an energy term $-h_p(\sum_{t\in\mathcal{S}_T.\backslash p}x_t - \sum_{k\in\mathcal{S}_K.\backslash p}x_k + x'_p - |\mathcal{S}_T.\backslash p|-  \mathbb{I}_{p\in\mathcal{S}_K}+\epsilon)$, where $|\mathcal{S}_T.\backslash p|$ is the size of set $\mathcal{S}_T$ after $p$ being removed. $x'_p=-x_p$ and $\mathbb{I}_{p\in\mathcal{S}_K}=0$ if $p \in \mathcal{S}_T$, otherwise $x'_p = x_p$ and $\mathbb{I}_{p\in\mathcal{S}_K}=1$.
\end{itemize}
Combine the energy terms above together we have the energy function of RBM with $T+K+1$ hidden units as in \eqref{sdnf_rbm}.
%
\end{proof}
In the case where $p$ is the last to be eliminated we do not need a hidden unit to represent an energy term. Let us consider the case $p\in \mathcal{S}_T$ then the energy term $-h_p(-x_p+\epsilon)$ can be replaced by $-(1-x_p)\epsilon$. This is possible because in order to minimise the energy, $h_p=1$ if and only if $x_p=0$, or in formal expression $h_p=1-x_p$. Therefore, $-h_p(-x_p+\epsilon) = -(1-x_p)(-x_p+\epsilon)=-(-x_p +\epsilon+x_p^2-x_p\epsilon) = -(1-x_p)\epsilon$, because $x_p=x_p^2$. Similarly, if $p\in \mathcal{S}_K$ then the energy term $-h_p(x_p-1+\epsilon)$ can be replaced by $-x_p\epsilon$. The final RBM for a logical implication in \eqref{horn} only needs $K+T$ hidden units.

\subsection{Comparison to Penalty Logic and RBMs as Universal Approximator}
\label{vspenalty}
\subsubsection{Penalty Logic}
The most related work to ours is Penalty logic \cite{Pinkas_1995}, which shows the representation of propositional formulas in symmetric connectionist networks, e.g. Hopfield networks and Boltzmann machines. In the first step Penalty logic adds more hidden variables to reduce a formula $\varphi$ to a conjunction of sub-formulas $\bAnd_i \varphi_i$, each of at most three variables. This ``naming'' step makes the conversion to energy function easier. However, some of the cubic terms in the energy function may consist of hidden variables and therefore we cannot convert $\varphi$ into an RBM. For example, a negative cubic term $-h_1xy$ of a high-order BM will be transformed into the quadratic term $-h_2h_1 - h_2x -h_2y + 5h_2$, with $-h_2h_1$ forms a connection between two hidden units. 

Let us take the logical implication as a case in point to compare Penalty logic and our theory. Without using ``naming'', the high-order energy of a logical implication \eqref{horn}, according to Penalty logic, is:
\begin{equation}
\label{bm_pen}
E = \prod_{t\in \mathcal{S}_T}x_t\prod_{k\in\mathcal{S}_K}(1-x_k) (1-y)
\end{equation}

In order to convert this high-order energy to a quadratic form of an RBM we have to expand \eqref{bm_pen} to obtain a sum of products. This would result in $\sum_{k=0}^{K+1} \binom{K+1}{k}=2^{K+1}$ energy terms having the orders ranging from $T$ to $T+K+1$. Also we need to note that, to reduce a positive term of higher-order energy to quadratic terms, the number of hidden variables to be added is proportional to the number of variables in that term.  Therefore, to convert a long formula the number of hidden variables needed will be exponentially large. Different from that, if we use SDNF then we will only need $T+K$ hidden variables for an RBM, as shown in the previous section. 

If $\mathcal{S}_K$ is empty, i.e. its size is zero $K=0$, then the logical implications $\varphi$  in \eqref{horn} becomes a Horn clause. Apply Theorem 3.2 in Penalty logic \cite{Pinkas_1995} the higher-order energy function \eqref{bm_pen} can be converted into a quadratic form of an RBM as:
{
\begin{equation*}
\begin{aligned}
&\En^{penalty}_{K=0} = -2h_y(\sum_{t\in \mathcal{S}_T}x_t +y -T+0.5)\\
&-\sum_{p\in\mathcal{S}_T}2h_{p}(\sum_{t\in \mathcal{S}_T.\backslash p}x_t - x_p-|\mathcal{S}_T.\backslash p| +0.5) + x_T
\end{aligned}
\end{equation*}
}
Meanwhile, from \eqref{sdnf_rbm} the energy function of an RBM using SDNF approach is:
{
\begin{equation*}
\begin{aligned}
&\En^{sdnf}_{K=0} = -h_y(\sum_{t\in\mathcal{S}_T}x_t +y -T-1+\epsilon)\\
&-\sum_{p\in\mathcal{S}_T}h_{p}(\sum_{t\in\mathcal{S}_T.\backslash p}x_t - x_{p}- |\mathcal{S}_T.\backslash p| +\epsilon ) - (1-x_T)\epsilon
\end{aligned}
\end{equation*}
}
We can see that if $\epsilon=0.5$ then $\En^{penalty}_{K=0} = 2*\En^{sdnf}_{K=0}+1$. In this case, the Penalty logic and SDNF can construct similar RBMs.
\subsubsection{RBMs as Universal Approximator}
From a statistical point of view, representing a formula in an RBM can be seen as approximating an uniform distribution over all preferred models. In \cite{LeRoux_2008}, it shows that any discrete distribution can be exactly presented by an RBM with $M+1$ hidden units where $M$ is the number of input vectors whose probability is not 0. In the case of logical implications $M=2^{K+T+1}-1$. This is similar to FDNF where only one conjunctive clause holds given a preferred model. However, in the case of SDNF while the universal approximators utilise all possible modes from the uniform distribution representing a formula our work focuses on transformation of that formula to construct a more compact RBM.
\begin{example}
Given a logical implication $\pr{y}\lif \pr{x}_1\fzand \neg \pr{x}_2 \fzand \neg \pr{x}_3$ we will show how different RBMs can be constructed by Penalty logic, RBMs as universal approximator and our approach. 

In Penalty logic, the first step is to create a higher-order energy function: $E=x_1(1-x_2)(1-x_3)(1-y)$. After that, by applying Theorem 3.2 in \cite{Pinkas_1995} this function can be reduced to a function of lower order until it has a quadratic form of an RBM. The result is an RBM with $4$ hidden units.
{
\begin{equation*}
\begin{aligned}
E=&-h_1(2x_1+2x_2+2x_3+2y-7)\\
  &-h_2(2x_1+2x_2-2x_3-3)\\
  &-h_3(2x_1+2y-2x_2-3)\\
  &-h_4(2x_1+2x_3-2y-3) + x_1
\end{aligned}
\end{equation*}
}
We can use Theorem 2.4 in \cite{LeRoux_2008} to construct an RBM as a universal approximator of the uniform distribution over the preferred models. In this case, from each sample (preferred model), e.g $[\pr{x}_1=True,\pr{x}_2=True,\pr{x}_3=False,\pr{y}=True]$ we create an input vector $\vt{v} = [1,1,0,1]$ with that we add to an RBM a hidden unit with a weight vector $\vt{w}=\vt{v}-\frac{1}{2} = [\frac{1}{2},\frac{1}{2},-\frac{1}{2},\frac{1}{2}]$ and a bias $c=-\vt{w}^\top\vt{v}+\lambda = -\frac{3}{2}+\lambda$ with $\lambda \in \mathbb{R}$. The result is an RBM with $15$ hidden units (we omitted the hidden unit whose weights and bias are zeros).

In the a case of our approach where SDNF is FDNF then using Theorem \ref{theorem:prop_rbm} we will construct a RBM with similar structure but different weights and biases. Now we detail how another RBM is constructed from Theorem \ref{theorem:horn}. Note that the steps below are for explanation purpose. In practice, RBMs can be directly constructed from the formula. First we transform $\pr{y}\lif \pr{x}_1\fzand \neg \pr{x}_2 \fzand \neg \pr{x}_3 \equiv (\pr{y} \fzand \pr{x}_1 \fzand \neg \pr{x}_2 \fzand \neg \pr{x}_3) \fzor  (\neg \pr{x}_1\fzor\pr{x}_2 \fzor \pr{x}_3)$. Then we gradually eliminating variables from the disjunctive clause. This can be done in any order but in this example we start from $\pr{x}_3$, $\pr{x}_2$, to $\pr{x}_1$ ($p= 3,2,1$). Here, $\mathcal{S}_T = \{1\}$ and $\mathcal{S}_K=\{2,3\}$. We have: 
{
\begin{equation*}
\begin{aligned}
(\neg \pr{x}_1\fzor\pr{x}_2 \fzor \pr{x}_3) &\equiv (\pr{x}_1 \fzand \neg \pr{x}_2 \fzand {\color{red}\pr{x}_3})\fzor (\neg \pr{x}_1 \fzor \pr{x}_2)\\
&\equiv
(\pr{x}_1 \fzand \neg \pr{x}_2 \fzand {\color{red}\pr{x}_3})\fzor (\pr{x}_1 \fzand {\color{red}\pr{x}_2})\fzor (\neg \pr{x}_1)
\end{aligned}
\end{equation*}
}
Now we construct an energy terms for each conjunctive clause:
{
\begin{equation*}
\begin{aligned}
(\pr{y} \fzand \pr{x}_1 \fzand \neg \pr{x}_2 \fzand \neg \pr{x}_3) &: e_y = -h_y(x_1-x_2-x_3+y-2+\epsilon)\\
(\pr{x}_1 \fzand \neg \pr{x}_2 \fzand {\color{red}\pr{x}_3}) &: e_3 = -h_3(x_1 -x_2 + x_3 - 2+\epsilon)\\
(\pr{x}_1 \fzand {\color{red}\pr{x}_2})&: e_2 = -h_2(x_1+x_2-2+\epsilon)\\
(\neg \pr{x}_1) &: e_1=(1-x_1)\epsilon
\end{aligned}
\end{equation*}
}
Together, they form the energy of an RBM with $3$ hidden units. 
\end{example}
\section{Confidence Rule Inductive Logic Programming}
In this section, we employ Theorem \ref{theorem:horn} to encode formulas in a knowledge base. Once the formulas are encoded, we can tune the knowledge by learning the parameters called {\it confidence values} associated with the conjunctive clauses, and statistical inference can  be employed for reasoning.

\subsection{Encoding}
\begin{figure}[ht]
\centering
\includegraphics[width=0.15\textwidth]{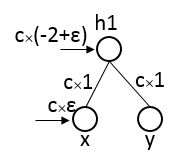}
\caption{An RBM represents $\pr{y}\lif \pr{x}$.}
\label{exp_rbm}
\end{figure}

By applying Theorem \ref{theorem:horn} and Proposition \ref{prop:rbm_lprogram} we construct a set of unique conjunctive clauses from the logical implications in a knowledge base. After that, each clause is associated with a non-negative real value named {\it confidence value} $c$. Encoding of such weighted clauses can be done as follows: first, we encode the clauses to an RBM, then we multiply the newly created parameters (connection weights and biases) with $c$.  For example, a formula $\pr{y}\lif \pr{x}$ with confidence value $c$ is encoded in an RBM as in Figure \ref{exp_rbm}.
After encoding, we can add more hidden units. Practically, this would help neural-symbolic integration models perform better \cite{Towel_1994,Garcez_1999}. 

\subsection{Learning}
Once the formulas are encoded into an RBM, we can now tune the background theory by training the model on an actual dataset to update the confidence values. It is expected that the learning process can generalise the rules better while preserving the knowledge in symbolic form. We can train the network by minimising the combined negative log-likelihood, similar to hybrid learning for RBM in \cite{Larochelle_2008}, as:   
\begin{equation}
\label{learning}
\sum_{\vt{x},y} (-\alpha \log p(\vt{x},y) - \beta \log p(y|\vt{x}) )
\end{equation}
where $\alpha$ and $\beta$ are real positive values indicating preference of minimising generative cost or discriminative cost respectively.

\subsection{Inference}
According to Proposition \ref{prop:gibbs_sat}, inference can be done by clamping the variables which have been assigned with truth values then iteratively inferring the hidden variables and the unassigned ones until the RBM converges to a stationary state. We can also infer the truth value of a variable by using a conditional distribution, i.e. from $p(y|\vt{x}) = p(y|\{x_t,x_k|t \in \mathcal{S}_T;k \in \mathcal{S}_K\}$. The advantage of this method is that if all $x_t$, $x_k$ are known then the distribution is tractable \cite{Larochelle_2008,Cherla_2017}. In this case the inference only takes one step, i.e. $y=1$ if $p(y=1|\vt{x})\geq0.5$ and otherwise.
\section{Experiments}
\label{exp}

\subsection{XOR \& Car Evaluation}
First, we investigate how a trained RBMs can approximately represent symbolic knowledge.
We train an RBM with four hidden units on the truth table of XOR ($(\pr{x}\oplus\pr{y})\fziff \pr{z}$). After training we approximate conjunctive clauses from the parameters of RBM by searching for ones that have minimum Euclidean distance to the column vectors in weight matrix. For example, a column vector $[6.2166,-6.7347,6.3059]^\top$ is closest to $6.419\times[1,-1,1]^T$ which represents a conjunctive clause: $\pr{x} \fzand \neg \pr{y} \fzand \pr{z}$ with confidence value $c=6.419$. 
{
\begin{equation*} 
W = \begin{bmatrix} -7.0283 & 5.6875 & -6.7200 & 6.2166 \\[0.3em]
                    -6.8593 & 6.0078 & 6.2008 &-6.7347 \\[0.3em] 
                   -6.9774  & -6.5855 & 6.0395 & 6.3059 \end{bmatrix}
\end{equation*}
}
{
\begin{equation*}
b = [5.7909, -6.3370, -6.3478, -6.2275]^\top
\end{equation*}
} The weight matrix and biases above resemble four Confidence
rules that represent XOR. 
\begin{figure}[ht]
\centering
\includegraphics[width=0.4\textwidth]{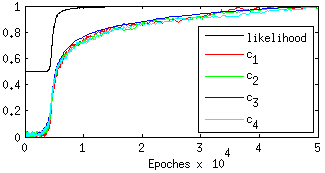}
\caption{The likelihood and approximate confidence values from RBM trained on XOR. The confidence values are scaled to $[0,1]$.}
\label{fig:cv_track}
\end{figure}
 If we run the training for long time, such confidence values are increasing
as the optimised function is approaching a local/global minimum.

Because the SDNF form of XOR is a FDNF  we cannot observe the indication of variable elimination mentioned in Theorem \ref{theorem:horn}  where more compact conjunctive clauses can be obtained. Therefore, we also train RBMs on Car Evaluation dataset \footnote{\url{https://archive.ics.uci.edu/ml/datasets/car+evaluation}} and extract  each conjunctive clause from a column vector to which it  has the shortest Euclidean distance. Here, many clauses can be obtained with some of variables being pruned out. We measure the quality of a clause using the reliability ratio, denoted as $rr = \frac{\text{\# of data samples that satisfy the clause}}{\text{\# of data samples that violate the clause}}$. A data sample is seen as a preferred model which satisfies a clause if both of them draw the same conclusion on whether a car is unacceptable or acceptable or good or very good given its  prices and technical specifications. Otherwise we call this as  a violation. We found that the shorter clauses with high confidence values are also more reliable. In Table \ref{tab:car_eval} we show five of the shortest clauses and the reliability ratios.
\begin{table}
\centering
{
	\begin{tabular}{lc}
	\hline
	clauses & rr \\
	\hline	
	{\tiny$1.62: \pr{unacc} \fzand \pr{safe\_l}$}  & {\scriptsize 576/0}\\
    {\tiny$1.47: \pr{unacc} \fzand \pr{2\_seats}$}  & {\scriptsize 576/0}\\
    { \tiny$1.33: \pr{acc} \fzand \pr{buy\_h} \fzand \pr{maint\_h} \fzand \pr{4\_seats} \fzand \pr{safe\_h}$} & {\scriptsize 108/0} \\
    {\tiny $0.74: \pr{good} \fzand \pr{buy\_l} \fzand \pr{maint\_l} \fzand \pr{4\_seats} \fzand \pr{big\_boot} \fzand \pr{safe\_m}$} & {\scriptsize 12/0}\\
	{\tiny $0.73: \pr{vgood} \fzand \pr{buy\_l} \fzand \pr{maint\_m} \fzand 4\_seats \fzand \pr{big\_boot} \fzand \pr{safe\_h}$}& {\scriptsize 4/0}\\
	\hline
	\hline
	\end{tabular}
	\caption{Reliability of extracted clauses. $\pr{unacc}$,$\pr{acc}$,$\pr{good}$,$\pr{vgood}$ are propositions for evaluation of a car: {\it unacceptable}, {\it acceptable}, {\it good}, {\it very good}; $\pr{safe\_l}$, $\pr{safe\_m}$, $\pr{safe\_h}$ are the safety: {\it low}, {\it medium}, {\it high}; $\pr{2\_seats}$, $\pr{4\_seats}$ are the number of seats; $\pr{buy\_l}$, $\pr{buy\_h}$ are the buying cost: {\it high}, {\it low};$\pr{maint\_l}$, $\pr{maint\_m}$, $\pr{maint\_h}$ are the maintenance cost: {\it high}, {\it medium}, {\it low}.}  
	\label{tab:car_eval}
}
\end{table}

\subsection{DNA Promoter}
The DNA promoter
dataset\footnote{\url{http://archive.ics.uci.edu/ml/datasets/Molecular+Biology+(Promoter+Gene+Sequences)}}
has 106 examples, each consists of a sequence of $57$ nucleotides from
position $−50$ to $+7$ in the DNA. Each nucleotide has a discrete
value $n \in \{a, t, g, c\}$. The data includes 53 examples of gene
promoters and 53 examples which are not gene promoters. Let us use a
literal $\pr{n}_p$ to denote that a nucleotide at position $p$ has a
value $n$. For example, $\pr{a}_{1}$, $\pr{t}_2$, $\pr{g}_3$,
$\pr{c}_4$ indicate that the nucleotides at positions $1$, $2$, $3$
and $4$ are $a$, $t$, $g$, $c$ respectively.  The background theory is
composed of $14$ rules in which besides the literals about the
nucleotides ( $\pr{n}_p$) and the promoters ($\pr{promoter}$,
$\neg \pr{promoter}$) there exist three intermediate literals
($\pr{minus}_{10}$, $\pr{minus}_{35}$, $\pr{conformation}$).  We apply
syllogism to eliminate the intermediate literals to make the rules easier to encode to an RBM. For examples:

\begin{equation*}
\begin{aligned}
&\pr{minus}_{10} \lif  \pr{t}_{-12} \fzand \pr{a}_{-11} \fzand \pr{t}_{-7}\\
&\pr{minus}_{35} \lif  \pr{t}_{-36} \fzand \pr{t}_{-35} \fzand \pr{g}_{-34} \fzand \pr{a}_{-33} \fzand \pr{c}_{-32}\\
&\pr{conformation} \lif \pr{a}_{-45} \fzand \pr{a}_{-44} \fzand \pr{a}_{-41}\\
&\pr{contact} \lif \pr{minus}_{35} \fzand \pr{minus}_{10}\\
&\pr{promoter} \lif \pr{contact} \fzand \pr{conformation}\\
\hline
&\pr{promoter} \lif \pr{t}_{-12} \fzand \pr{a}_{-11} \fzand \pr{t}_{-7} \fzand \pr{t}_{-36} \fzand \pr{t}_{-35} \fzand \pr{g}_{-34}\\
& \hskip 1.8cm\fzand \pr{a}_{-33} \fzand \pr{c}_{-32} \fzand \pr{a}_{-45} \fzand \pr{a}_{-44} \fzand \pr{a}_{-41}
\end{aligned}
\end{equation*}
We then encode the final rules to an RBM to build
a CRILP as detailed earlier. In particular, for each literal $\pr{n}_p$ we create a
visible unit and also two target units were also added, one for
$\pr{promoter}$ and the other for $\neg \pr{promoter}$. After that 10 more hidden units with random weights are added. In this
experiment leave-one-out cross validation is used for CRILP which
achieves $93.16\%$ accuracy. For comparison, C-IL$^2$P and CILP++
achieve $92.41\%$ and $92.48\%$ accuracy respectively.
\begin{table*}[ht]
\centering
{
\begin{tabular}{|l|c|c|c|c|c|}
\hline
            & { Aleph}    & {CILP++}     & {CRILP$_{h0}$} & {CRILP$_{h50}$} & RBM$_{h100}$ \\
 \hline
 {\small Muta}       & 80.85    & 91.70      & 95.17	 & \textbf{96.28} & 95.55\\
 
 {\small krk}        & 99.60    & 98.42      & 97.10   & \textbf{99.80} & 99.70\\
 {\small uw-cse}     & 84.91    & 70.01      & 87.57 & \textbf{89.43} & 89.14\\
 {\small alz-amine}  & 78.71    & 78.99      & 72.29   & 78.25 & \textbf{79.13}\\
 {\small alz-acetyl} & \textbf{69.46}    & 65.47      & 66.97   & 66.82 & 62.93\\
 {\small alz-memory} & 68.57    & 60.44      & 64.71   & \textbf{71.84} & 68.54\\
 {\small alz-toxic}  & 80.50    & 81.73      & 81.35   & \textbf{84.95} & 82.71\\
 \hline
 \hline
\end{tabular}
}
\caption{Evaluation of Aleph, CILP++ and CRILP on 7 datasets}
\label{tab:ilp}
\end{table*}

\subsection{Inductive Logic Programming}
We also tested our approaches of encoding, learning and inference for the inductive programming task. In this experiment, first we extract bottom clauses from  examples and background knowledge. Then we encode the bottom clauses  into RBMs and tune the confidence values using a small number of preferred models, one from each clause. 

We compare our system with Aleph \cite{aleph} and CILP++ \cite{Franca_2014}. Aleph is a state-of-the-art inductive logic programming system built purely on symbols. It performs hypothesis search for generalised formulas from examples and background theory. CILP++ is a state-of-the-art neural-symbolic logic inductive programming system based on feed-forward/recurrent neural networks. CILP++ is developed by extending the CILP system \cite{Garcez_1999}. Both Aleph and CILP++ are based upon {\it bottom clause propositionalisation} so we also build our system on that. First, we have the bottom clauses extracted from the examples and background theory  by the same procedure in \cite{Franca_2014}.  Then the bottom clauses will be used for encoding and tuning. For tuning (training confidence values), we use one preferred model from each clause. In particular, for  example, from clause $y \lif \pr{x}_1 \fzand \neg \pr{x}_2$ we take the preferred model $y=1,x_1=1,x_2=0$ which set both sides of the symbol $\lif$ to $True$ as a training sample vector $[1,1,0]$. 

We carry out the experiments on 7 datasets: Mutagenesis \cite{Srinivasan_1994}, KRK \cite{Bain_1994}, UW-CSE \cite{Richardson_2006}, Amine, Acetyl, Memory and Toxic.  The last four datasets are from Alzheimer benchmark \cite{King_1995}.
 We test two different CRILPs. One is constructed only from bottom clauses (all), without any additional hidden units. We call it CRILP$_{h0}$. The other is CRILP$_{h50}$, is constructed from a subset of bottom clauses with $50$ more hidden units added. In particular, we use $2.5\%$ bottom clauses from  the Mutagenesis and KRK datasets. For the larger datasets from UW-CSE and Alzheimer benchmark we use $10\%$ of bottom clauses. For completeness, we also compare our CRILPs with RBMs having $100$ hidden units fully connected to the visible units. This number of hidden units is to make sure that the RBMs have more parameters than our CRILPs in all cases.

The experimental results are shown in Table \ref{tab:ilp} using 10 fold cross validation for all datasets, except the UW-CSE which has 5-folds. For learning, as shown in \eqref{learning}, we set $\beta=1$ and select $\alpha$ from either $0$ or $0.01$. For inference, we use conditional distribution $p(y|\vt{x})$. The results of Aleph and CILP++ are copied from \cite{Franca_2014}. It can be seen  that CRILP$_{h50}$ has the best performance in 5 out of 7 cases. In {\it alz-acetyl} dataset Aleph is better than the others model used in this evaluation, and RBM$_{h100}$ is ranked first  for {\it alz-amine} dataset. The experimental results of CRILP$_{h0}$ show that the encoded prior knowledge can perform well without adding more hidden units. Also, we can see that improvement can be achieved by adding more hidden units, as shown in the results of CRILP$_{h50}$.  
\section{Conclusion}
\label{sec:discussion}
This paper proposes a method to unify propositional logic and neural
networks when restricted Boltzmann machines are employed. First, we
show how to represent a propositional logic program in an unsupervised
energy-based connectionist network efficiently. Second, we demonstrate
a potential application of this work for integration of neural
networks and symbolic formulas where background knowledge is encoded
into RBMs. The results from this work can set up a theoretical
foundation for further exploration of relationships between
unsupervised neural networks and symbolic representation and
reasoning. 
\bibliographystyle{aaai}
\bibliography{biblio}
\end{document}